\documentclass[letterpaper, 10 pt, conference]{ieeeconf} 

\IEEEoverridecommandlockouts

\usepackage{etextools}
\usepackage{amssymb}
\usepackage{float}
\usepackage{makeidx}
\usepackage{amsmath}
\usepackage{bbm}
\usepackage{epsfig}
\usepackage{epsf}
\usepackage{psfrag}
\usepackage{verbatim}
\usepackage{color}
\usepackage{multirow}
\usepackage{tabularx}
\usepackage{booktabs}
\usepackage[tight,footnotesize]{subfigure}
\usepackage{array}
\usepackage{soul}
\usepackage{footnote}
\usepackage{cite}
\usepackage{dblfloatfix}

\usepackage{color, colortbl}
\usepackage[colorlinks,bookmarksnumbered,citecolor=orange,urlcolor=orange]{hyperref}
\usepackage{graphicx}
\graphicspath{{Figures}}
\DeclareGraphicsExtensions{.pdf,.png}
\usepackage{bigstrut}
\usepackage[english]{babel}

\usepackage{csquotes}

\usepackage[T1]{fontenc}
\usepackage{algorithm, setspace}
\usepackage{algpseudocode}
\usepackage{url}
\usepackage{multirow}
\usepackage{float}
\usepackage{xcolor}
\usepackage{hyperref}
\usepackage{booktabs}
 \hypersetup{
     colorlinks=true,
     linkcolor=orange,
     filecolor=orange,
     citecolor=orange,      
     urlcolor=orange,
     }

\newcommand{\p}[1]{\smallskip \noindent \textbf{{#1}.}}

\newcommand{\eq}[1]{Equation~(\ref{eq:#1})}
\newcommand{\fig}[1]{Figure~\ref{fig:#1}}

\usepackage{balance}

\title{\LARGE

RISO: Combining Rigid Grippers with Soft Switchable Adhesives

}

\author{Shaunak A. Mehta, Yeunhee Kim, Joshua Hoegerman, Michael D. Bartlett, and Dylan P. Losey
\thanks{This work is supported by NSF Grant $\#2205241$.}
\thanks{The authors are members of the Department of Mechanical Engineering, Virginia Tech, Blacksburg, VA 24061.
\newline
{e-mails: \texttt{\{mehtashaunak, yhkim39, jhoegerm, mbartlett, losey\}@vt.edu}}}
}

\begin{document}
\maketitle

\begin{abstract}
Robot arms that assist humans should be able to pick up, move, and release everyday objects. Today's assistive robot arms use \textit{rigid grippers} to pinch items between fingers; while these rigid grippers are well suited for large and heavy objects, they often struggle to grasp small, numerous, or delicate items (such as foods). \textit{Soft grippers} cover the opposite end of the spectrum; these grippers use adhesives or change shape to wrap around small and irregular items, but cannot exert the large forces needed to manipulate heavy objects. In this paper we introduce RIgid-SOft (RISO) grippers that combine switchable soft adhesives with standard rigid mechanisms to enable a diverse range of robotic grasping. We develop RISO grippers by leveraging a novel class of soft materials that change adhesion force in real-time through pneumatically controlled shape and rigidity tuning. By mounting these soft adhesives on the bottom of rigid fingers, we create a gripper that can interact with objects using either purely rigid grasps (pinching the object) or purely soft grasps (adhering to the object). This increased capability requires additional decision making, and we therefore formulate a shared control approach that partially automates the motion of the robot arm. In practice, this controller aligns the RISO gripper while inferring which object the human wants to grasp and how the human wants to grasp that item. Our user study demonstrates that RISO grippers can pick up, move, and release household items from existing datasets, and that the system performs grasps more successfully and efficiently when sharing control between the human and robot. See videos here: \url{https://youtu.be/5uLUkBYcnwg}
\end{abstract}


\section{Introduction}

\begin{figure}[t]
    \begin{center}
        \includegraphics[width=1.0\columnwidth]{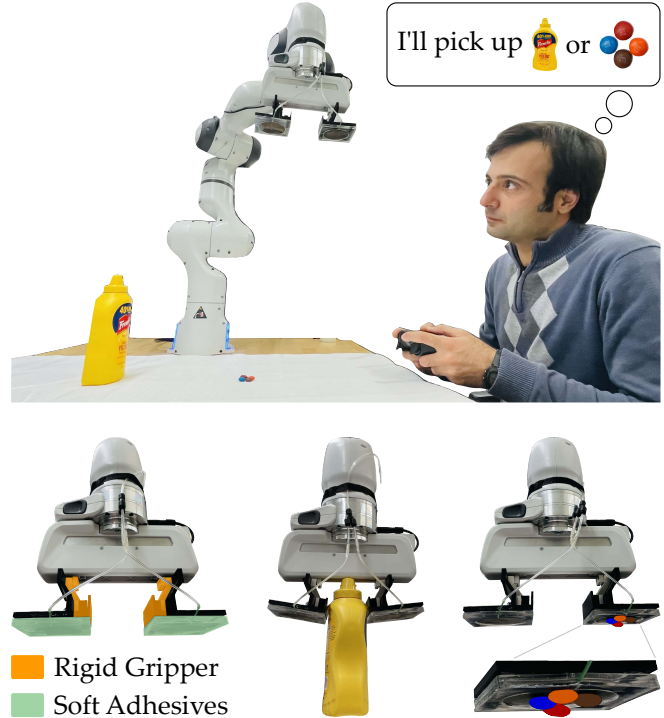}
        \caption{Human controlling an assistive robot arm and RIgid-SOft (RISO) gripper. The RISO gripper is formed by mounting soft, switchable adhesives to the bottom of a rigid, industry-standard, parallel mechanism. Using RISO grippers robot arms can pick up large, rigid items (e.g., a jar of mustard) as well as small, numerous, and fragile objects (e.g., a pile of candy).}
        \label{fig:front}
    \end{center}
    \vspace{-2.0em}
\end{figure}

\begin{figure*}[t]
    \begin{center}
        \includegraphics[width=2.0\columnwidth]{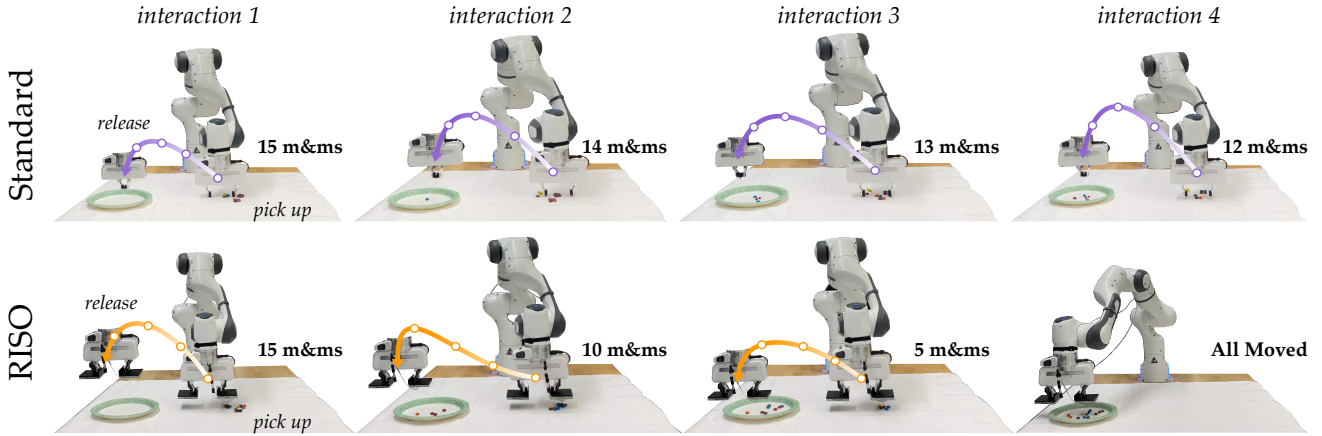}
        \caption{Moving small, numerous objects with an industry-standard rigid gripper \cite{frankagripper} and our proposed RISO gripper. The robot was tasked with carrying $15$ M\&M's from piles on the right side of the table to a plate on the left side of the table. (Top) Using the standard rigid gripper the robot was only able to move one item per interaction. (Bottom) By contrast, the RISO gripper could pick up, hold, and release multiple small candies during each interaction.}
        \label{fig:example}
    \end{center}
    \vspace{-2.0em}
\end{figure*}

Robot arms should be able to grasp everyday objects (see \fig{front}). For instance, consider the over $1$ million American adults living with physical disabilities who need assistance during activities of daily living \cite{taylor2018americans}. Wheelchair-mounted robot arms should offer these adults an avenue to pick up, move, and release household items. Today's assistive robot arms apply \textit{rigid grippers} --- such as parallel, multi-fingered grippers --- to provide a wide range of precise grasping forces \cite{beaudoin2019long, jain2016grasp, frankagripper, robotiq}. If the desired object is large and rigid (e.g., a jar of mustard), these rigid grippers can squeeze and hold that object. But what about objects that are small, numerous, or delicate (e.g., a pile of candy)? For these everyday objects rigid grippers fall short: at best, the gripper can pinch and hold one item at a time, and if the object is too small or soft, then it may slip through the rigid gripper or suffer damage during the grasping process (see \fig{example}).

Recent works address these shortcomings by creating \textit{soft grippers}. Soft grippers such as gecko-inspired adhesives \cite{glick2018soft, song2014geckogripper, tian2019gecko, hao2020multimodal}, granular jamming \cite{brown2010universal}, and electroadhesion \cite{shintake2016versatile, alizadehyazdi2020electrostatic} complete grasps by leveraging adhesion or conforming to the object's shape. However, these grippers also come with a trade-off. On the one hand, they enable robot arms to pick up small, fragile, and irregular objects that may not be possible with rigid grippers; on the other hand, soft grippers are often unable to grasp large, heavy items or objects with rough surfaces that are suitable for rigid grippers \cite{shintake2018soft}. Put another way, rigid and soft grippers have separate grasping domains.

When a robot arm uses grippers that are either purely rigid or purely soft, it fundamentally limits the types of objects that robot can manipulate. Instead of applying a single gripper type, our insight is that:
\begin{center}\vspace{-0.3em}
\textit{Mounting \emph{soft} controllable adhesives on standard \emph{rigid} grippers enables a diverse range of robotic grasping.}
\vspace{-0.3em}
\end{center}
In this paper we introduce a novel \textbf{RI}gid-\textbf{SO}ft (\textbf{RISO}) gripper that integrates elements of both rigid and soft paradigms (see \fig{front}). RISO takes advantage of soft materials that can rapidly and repeatedly switch between low and high adhesion \cite{swift2020active}. We attach these \textit{switchable adhesives} to the bottom of a standard parallel gripper: when the robot moves to pick up an object, it can either (a) pinch the object using the rigid gripper, or (b) stick to the object using the soft adhesive. As a first step towards characterizing this unified gripping formulation, we here explore how humans and robot arms control the RISO gripper to pick up and release everyday objects with diverse sizes, weights, and shapes.

Overall, we make the following contributions:

\p{Creating RISO Grippers} We introduce RIgid-SOft (\textbf{RISO}) grippers by attaching a novel class of switchable adhesives to the bottom of industry-standard rigid mechanisms. Unlike other rigid-soft designs \cite{ham2018soft, park2018hybrid, tang2019development, wu2019novel, guo2020self, gafer2020quad, yang2020rigid, hussain2020design, li2022stiffness}, RISOs leverage adhesives to perform grasps that are purely rigid or purely soft.

\p{Sharing Gripper Control} When using a RISO gripper operators may grasp items with either the soft adhesives or the rigid mechanism. This results in additional decision making (e.g., choosing the correct grasp type or adhesive pressure). Accordingly, we formulate a shared control approach for RISO grippers that partially automates the grasping process while leaving the human in charge of key decisions.

\p{Conducting User Studies} We perform an in-person user study where a robot arm with a RISO gripper must pick up, move, and release a dataset of household objects. We compare the gripper's performance when it is fully teleoperated by the human and when it uses our shared control formalism to partially automate the grasping process.

\section{Related Work}

\noindent \textbf{Rigid Grippers.} Industry-standard rigid grippers such as Robotiq products \cite{robotiq}, the Franka Hand \cite{frankagripper}, and the JACO gripper \cite{jain2016grasp} use parallel mechanisms to pinch a target object between two or more fingers. These grippers are capable of large and precise forces; for instance, the Franka Hand used in our experiments can apply a continuous force of up to $70$~N across a stroke of $80$~mm. But while these grippers are suitable for large and heavy objects, they cannot easily grasp small, numerous, or irregularly shaped items (see \fig{example}).

\p{Soft Grippers} Soft grippers seek to address the shortcomings of their rigid counterparts. There are a variety of soft gripper designs \cite{shintake2018soft, brown2010universal}: most relevant are approaches that leverage \textit{adhesives}. This includes gecko-inspired adhesives \cite{glick2018soft, song2014geckogripper, tian2019gecko, hao2020multimodal}, electrostatic adhesives \cite{alizadehyazdi2020electrostatic, shintake2016versatile}, dry adhesives \cite{hu2021soft}, and thermal adhesives \cite{coulson2022versatile}. Across each of these approaches the adhesive is combined with soft materials to create a overall compliant gripper; put another way, the robot never uses rigid components to interact with target objects. Although soft designs improve the robot's ability to grip small and irregular objects, the lack of rigid structure fundamentally limits the gripper's capacity to exert large forces \cite{shintake2018soft}.

\p{Rigid-Soft Grippers} Our hypothesis is that the combination of rigid and soft elements opens the door to diverse object manipulation. Recent works have moved towards \textit{rigid and soft} grippers by integrating both elements into the robot's fingers \cite{ham2018soft, park2018hybrid, tang2019development, wu2019novel, guo2020self, gafer2020quad, yang2020rigid, hussain2020design, li2022stiffness}. We note that these existing rigid and soft gripper designs \textit{do not} leverage adhesives. For example, Park \textit{et al.} \cite{park2018hybrid} intersperse rigid blocks throughout soft materials, while Guo \textit{et al.} \cite{guo2020self} vary a finger's stiffness by locking its rigid backbone in place. Gafer \textit{et al.} \cite{gafer2020quad} similarly connect rigid links with soft joints to make fingers that scoop-up items. Hussain \textit{et al.} \cite{hussain2020design} and Ham \textit{et al.} \cite{ham2018soft} use a tendon-driven mechanism to control the stiffness of the fingers, while Wu \textit{et al.} \cite{wu2019novel} integrate a jointed endoskeleton structure with soft robotic fingers to provide a higher gripping force. Overall, these state-of-the-art designs result in grasps that are both rigid and soft: the object is held by fingers that have intermixed rigid and compliant components. By contrast, RISO grippers can perform grasps that are \textit{purely rigid} (using industry-standard rigid grippers) or \textit{purely soft} (using adhesives placed at the tips of the rigid mechanism). This rigid high-level and soft low-level design is made possible by the use of a novel class of switchable adhesives.


\p{Switchable Adhesives} To create RISO grippers we leverage recent advances in adhesive materials \cite{swift2020active}. These adhesives are switchable: by activating a trigger (the input pneumatic pressure to a soft membrane) we control the level of adhesion to rapidly pick-up and release objects. Current research has focused on characterizing the minimum and maximum adhesion forces, as well as how quickly the material can switch between extremes \cite{croll2019switchable, tatari2018dynamically, frey2022octopus}. However, previous work has not applied these soft adhesives to robot grippers.
\section{RIgid-SOft (RISO) Grippers}

In this section we present our physics and control framework for RISO grippers. Overall, we form RISOs by mounting sheets of switchable adhesives to the fingers of a parallel rigid gripper (see \fig{adhesive}). When reaching for an object, the robot arm can either (a) adhere to the item by changing the adhesion of the soft materials, or (b) squeeze the item by actuating the rigid gripper. The rigid and soft elements are \textit{independent} --- the low-level soft materials do not interfere with the rigid grasp, and the high-level rigid fingers do not change the shape of the adhesives or wrap them around target objects. In Section~\ref{sec:adhesive} we first describe the fabrication process and the hardware setup and then explain the mechanics behind the switchable adhesives. Next, in Section~\ref{sec:control} we develop a shared control framework that partially automates RISO grasps by inferring which object the human wants to grasp and how the human wants to grasp that item.

\subsection{Switchable Adhesives} \label{sec:adhesive}

\begin{figure}[t]
    \begin{center}
        \includegraphics[width=1.0\columnwidth]{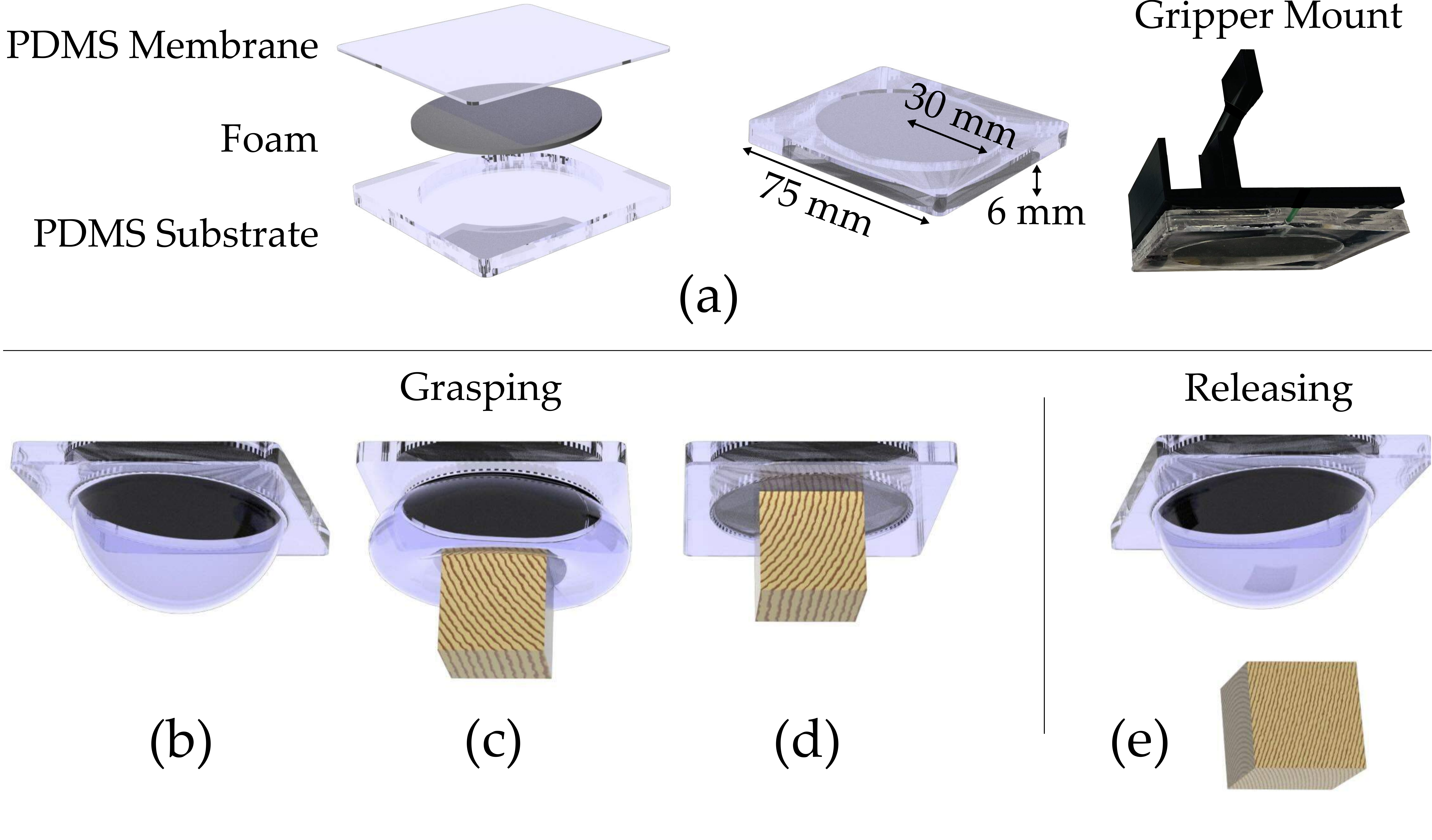}
        \caption{Soft switchable adhesives used in the RISO gripper. (a) Active membrane-foundation adhesives composed of a polydimethylsiloxane (PDMS) substrate, a foam layer, and then a PDMS membrane. We attach these adhesives to the bottom of the rigid gripper using $3$D printed mounts. (b -- e) Grasping and releasing an object. For grasping, negative chamber pressures ($P<0$) deflate the membrane and cause the object to adhere to the foam foundation. For releasing, positive pressures ($P>0$) inflate the membrane and decrease the contact area, reducing the adhesion force.}
        \label{fig:adhesive}
    \end{center}
    \vspace{-2.0em}
\end{figure}

\p{Structure and fabrication} 
Our switchable adhesives are an active membrane-foundation adhesives (AMFAs), which is reported in \cite{swift2020active}. 
This soft adhesive gripper consists of an elastomeric \textit{active membrane} composed of Polydimethylsiloxane (PDMS, Sylgard 184) substrate supported on a \textit{compliant foam foundation} which is embedded in a PDMS substrate.
The porous, compliant foundation is an open cell polyurethane foam (Poron Very Soft 20 pcf Microcellular Polyurethane, Rogers Corporation) that changes stiffness during pressure activation. This foam i) is compliant and enables intimate contact between the adhesive membrane and the target object, ii) is porous and quickly allows positive and negative pneumatic pressures to be supplied, iii) changes shape uniformly due to the near zero Poisson's ratio which assists with contact formation, and iv) can be used over many cycles for repeated use. Our resulting combination of an active membrane and compliant foam foundation forms a soft material that can apply a wide spectrum of programmable adhesion forces. When applying a positive pressure ($P>0$) the membrane inflates and adhesion is reduced; when applying a negative pressure ($P<0$) the membrane compresses, dynamically adjusting stiffness and resulting in a rapid and dramatic increase in output adhesion. Significantly, these active adhesives are \textit{switchable} in $\sim 0.1$ seconds and switch adhesion orders of magnitude faster than alternatives \cite{swift2020active}. Overall, the mechanical compliance, high adhesion range, and switchable nature of our soft adhesives makes them suitable for real-time object manipulation.

Substrates are fabricated using acrylic molds, where PDMS with a 10:1 base:curing agent ratio are cast and cured for 12 h at 40 $^{\circ}$C. 
Membranes are fabricated using PDMS with 15:1 base:curing agent ratio, and cured for 1 h at 80 $^{\circ}$C with a thickness of 160 $\mu$m. The thickness of the foam is nominally 1.6 mm, and the top surface of the foam is sanded with 80-grit sandpaper to prevent sticking to the membrane and is inserted between the substrate and the membrane (Figure \ref{fig:adhesive} (a)). To attach the membrane and substrate, we treated the surface of the substrate and membrane with plasma cleaning (30 s under 0.6 torr oxygen at medium RF level, PDC-001-HP, Harrick Plasma) and Silpoxy silicone adhesive, and pressed the treated surfaces together at room temperature for at least 12 h. After curing, a narrow rigid tube was embedded in the PDMS body for pressure control. The switchable adhesive is symmetric and the length of one side is 75 mm with a 30 mm radius active adhesive area, and an overall thickness of 6 mm, as shown in Figure \ref{fig:adhesive} (a). Additionally, the 3D-printed rigid part and switchable adhesives were attached using Silpoxy silicone adhesive. 

\begin{figure}[t]
    \begin{center}
        \includegraphics[width=1.0\columnwidth]{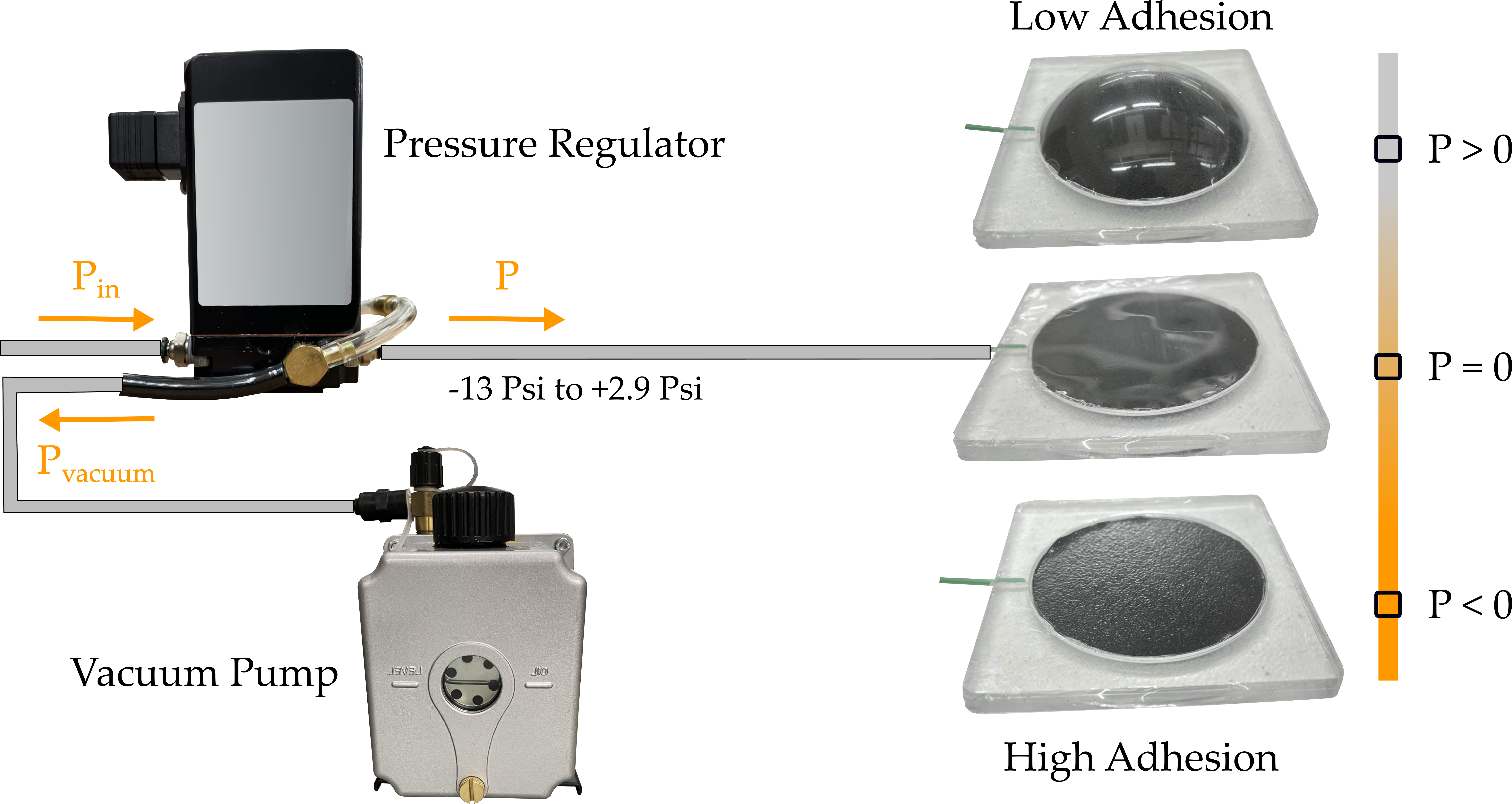}
        \caption{Pneumatic control setup for the RISO gripper's soft adhesives. (Left) The pressure regulator inlet ($P_{in}$) is connected to a supply pressure of $5$-$10$ Psi, and the exhaust is connected to a vacuum pump. The pressure $P$ of the output chamber ranges from $-13$ to $+2.9$ Psi. (Right) Adhesion force increases as we decrease $P$ from positive to negative values. The robot arm communicates with the pressure regulator to set $P$ in real-time.}
        \label{fig:setup}
    \end{center}
    \vspace{-2.0em}
\end{figure}

\p{Operation Principle} 
The operation of the switchable adhesives is controlled by a pneumatic control system (Figure \ref{fig:setup}). First, we inflate the membrane to increase the contact area between the membrane and the object (Figure \ref{fig:adhesive} (b)). After, we remote the manipulator to adhesive adhere to the object (Figure \ref{fig:adhesive} (c)). When the object is in contact with the AMFAs, we apply the internal chamber of AMFAs with the negative pressure, so that the membrane is tightly in contact with the foam (Figure \ref{fig:adhesive} (d)). When the negative pressure ($P < 0$) is applied, the membrane compresses the foam, and stiffness increases, resulting in enhanced adhesion that can grasp the object. Lastly, to release the object, we apply positive pressure to inflate the membrane (Figure \ref{fig:adhesive} (e)). When the positive pressure ($P > 0$) is applied, the membrane inflates and adhesion is reduced to enable switchable adhesion. 

\p{Mechanism of adhesion control}
Physical models for adhesive force capacity can guide the design and use of the switchable adhesives in the RISO. We refer to this range of achievable forces as the soft adhesive gripper's \textit{force capacity} $F_c$. Through a robust adhesion scaling theory we can quantify the force capacity\cite{bartlett2012looking,bartlett2013scaling}:
\vspace{-0.25em}\begin{equation} \label{eq:adh_1}
    F_c \propto \sqrt{G_c} \cdot \sqrt{A / C}
\vspace{-0.25em}\end{equation}
Here $A$ is the contact area between the gripper and the target object, $C$ is the gripper's compliance in the loading direction, and $G_c$ captures how 'adhesive' the interface between the gripper and the object is. More formally, $G_c$ is adhesive fracture energy. Equation~(\ref{eq:adh_1}) describes the range of forces our low-level adhesives can apply \textit{without assuming} any specific gripper geometry. 

To describe the adhesive force of the AMFA system, we consider that the compliance $C$ can be tuned through a change in pneumatic pressure $ P$, where the object is nominally $R$ in dimension on the contacting surface: 
\begin{equation}\label{force_pred}\tag{2}
F_{c} \propto \sqrt{\frac{G_{c} R^2}{C( P)}}
\end{equation}
This shows the dependence of $F_{c}$ on the applied pneumatic pressure and object size. 
\subsection{Shared Control} \label{sec:control}

Now that we have developed the physics behind our RISO gripper, our next step is to \textit{control} that gripper during human-robot interaction. This is challenging because the added capabilities of RISO grippers introduce additional decision making. Consider the example in \fig{front} where a human operator is teleoperating a robot arm to manipulate objects on a table. From the \textit{human's perspective}, operators need to be able to intuitively control the high-level, dexterous motion of the robot arm and the low-level, precise actions of the RISO gripper. From the \textit{robot's perspective}, the system must determine what object the human is currently trying to reach, how the human wants to grasp that item (i.e., using the rigid fingers or a soft adhesive), and how the robot should pick up that item (i.e., how much force or pressure should the RISO apply). In this section we formalize a shared control approach for assisting both the human and RISO robot with each of these decisions. Our proposed algorithm partially automates the motion of the robot arm to align the human's preferred gripper with their intended object; we then give the operator direct control over the precise grasp.

\p{Formulation} Recall that our RISO gripper is reaching for objects with either the rigid mechanism or one of the soft adhesives. Let $\mathcal{O}$ be the set of objects in the environment, and let $o_i \in \mathcal{O}$ be the pose of the $i$-th object. The robot tries to pick up these objects using grasp types $g \in \mathcal{G}$, where $\mathcal{G} = \{\text{rigid}, \text{soft}_1, \ldots, \text{soft}_n\}$ includes the rigid gripper and all of the $n$ soft adhesives. At the start of each interaction the robot does not know which item the person wants or how it should grasp that item. Let $o^* \in \mathcal{O}$ be the operator's desired object and let $g^* \in \mathcal{G}$ be their intended grasp type.

The human interacts with this system using a real-time teleoperation interface (e.g., a joystick). We give the human \textit{direct control} over the gripper's force and pressure, but \textit{share control} over the motion of the robot arm. Towards this end we define $s \in \mathcal{S}$ as the system state. Here $s$ includes $s_{\text{rigid}}$, the pose of the rigid gripper, $(s_{\text{soft}_1}, \ldots, s_{\text{soft}_n})$, the pose of each soft adhesive, $f$, the force applied by the rigid gripper, and $P$, the input pressure for the soft adhesives.

\smallskip

\noindent \textit{Direct Gripper Control.}
Users directly regulate the gripper force $f$ and pressure $P$. Specifically, humans input commands $\Delta f$ and $\Delta P$, and the RISO updates according to:
\begin{equation} \label{eq:C1}
    f^{t+1} = f^t + \Delta f^t, \quad P^{t+1} = P^t + \Delta P^t
\end{equation}
where $t$ is the current timestep.

\smallskip

\noindent \textit{Shared Arm Control.}
The human and robot share control over the motion of the robot arm (i.e., the robot's joint velocity). Let $a_{\mathcal{H}} \in \mathcal{A}$ be the human's commanded joint velocity, and let $a_{\mathcal{R}} \in \mathcal{A}$ be an autonomous action. We linearly blend these actions to find the arm's velocity $a$:
\begin{equation} \label{eq:blend}
    a = \alpha \cdot a_\mathcal{H} + (1 - \alpha) \cdot a_\mathcal{R}
\end{equation}
Here $\alpha \in [0, 1]$ is a design parameter that arbitrates control between the human and assistive robot \cite{losey2022learning, jain2019probabilistic, javdani2018shared}. When designers set $\alpha \rightarrow 1$, the human has full control over the robot arm, and when designers set $\alpha \rightarrow 0$, the robot increasingly automates its motion. Given joint velocity $a$ and the human's commanded changes in force and pressure, the system state $s$ transitions based on dynamics $T$: 
\begin{equation} \label{eq:dynamics}
    s^{t+1} = T(s^t, a^t, \Delta f^t, \Delta P^t)
\end{equation}
Overall, the goal of our shared control formulation is to select autonomous joint velocities $a_\mathcal{R}$ that (a) move the RISO gripper towards the human's intended object $o^*$, and (b) align the RISO gripper for the human's preferred grasp $g^*$. Imagine that the human in \fig{front} wants to pick up the candies using the right adhesive: an intelligent robot arm should move the RISO to align this adhesive directly above the desired candies. In what follows we describe how the robot infers both $o^*$ and $g^*$ in order to select assistive actions $a_\mathcal{R}$.

\p{Belief over Objects and Grasps} Belief $b$ captures the joint probability of an object and grasp given the previous system states and the human's actions in those states: 
\begin{equation}
    b^{t+1}(o, g) = P(o, g \mid s^{0:t}, a_\mathcal{H}^{0:t})
\end{equation}
Applying Bayes' theorem, and recognizing that the human's commands are conditionally independent \cite{javdani2018shared}, we reach:
\begin{equation} \label{eq:belief}
    b^{t+1}(o, g) \propto P(a_\mathcal{H}^t \mid s^t, o, g) \cdot b^t(o, g)
\end{equation}
The likelihood function $P(a_\mathcal{H}^t \mid s^t, o, g)$ captures the probability of the human commanding the robot arm to move with velocity $a_\mathcal{H}$ given that the system is in state $s$ and the human wants to pick up object $o$ with grasp $g$. We assume that the human takes actions to align the RISO gripper with their target object. Returning to our running example, if the operator wants to pick up candies with the soft adhesive, we anticipate the human will teleoperate the robot arm so that its soft adhesive is as close to the candies as possible. Consistent with prior work on behavioral economics and robotics \cite{luce2012individual, jeon2020reward}, we formally model the human as a nosily-optimal agent that seeks to minimize the distance between their desired object and their preferred gripper:
\begin{equation} \label{eq:likelihood}
    P(a_\mathcal{H} \mid s, o, g) \propto \exp{ \beta \Big( \|o, s_g\|^2 - \|o, s'_g\|^2 \Big)}
\end{equation}
Recall that $o \in \mathcal{O}$ is the pose of an object, and $s_g$ is the pose of gripper type $g$ (i.e., the pose of the rigid fingers or one of the soft adhesives). Here $s'_g$ is the next pose that gripper $g$ \textit{would have} if the robot arm takes action $a_\mathcal{H}$ and transitions according to \eq{dynamics}. The parameter $\beta \geq 0$ models how precisely the human controls the robot arm: as $\beta \rightarrow \infty$ we model the human as an increasingly optimal teleoperator, and learn more rapidly from each of the human's commands.

Intuitively, \eq{likelihood} asserts that an operator is likely to take actions that move their intended RISO gripper type to their target object. Combining \eq{likelihood} and \eq{belief}, the robot increases its confidence in objects $o$ and grasp types $g$ that match the human's commands.

\p{Autonomous Assistance} Now that we have an estimate of what the human wants, the robot can provide autonomous assistance to help perform that motion. Recall that $a_\mathcal{R}$ is the robot's autonomous arm velocity. We choose $a_\mathcal{R}$ to move towards the mean over the human's preferences:
\begin{equation} \label{eq:ar}
    a_\mathcal{R} = \sum_{o \in \mathcal{O}} \sum_{g \in \mathcal{G}} (o - s_g) \cdot b(o, g)
\end{equation}
As the robot updates its belief and becomes confident in the human's intent, \eq{ar} autonomously moves the end-effector so that the correct grasp type (rigid or soft) is aligned directly above the preferred object. In practice, we note that this approach is influenced by the \textit{prior} over objects and grasps, $b^0(o, g)$. Designers can tune this prior to reflect items that the human commonly reaches for, causing the robot to assist for these common grasps by default.

\p{Algorithm Summary} The goal of our shared autonomy approach is to help line up the robot arm so that humans can more easily leverage the RISO gripper. At each timestep the robot measures the human's input, $a_\mathcal{H}$, and uses \eq{belief} and \eq{likelihood} to infer which objects and grasps match the human's commands. The robot then selects autonomous action $a_\mathcal{R}$ using \eq{ar}. This action assists the robot towards likely grasps; the robot shares control between $a_\mathcal{H}$ and $a_\mathcal{R}$ using \eq{blend}. Overall, our algorithm has similarities to state-of-the-art approaches \cite{losey2022learning, jain2019probabilistic, javdani2018shared}. However, a key difference here is that the robot thinks about both the robot arm \textit{and the RISO gripper} when assisting the human.

\begin{figure*}[t]
    \begin{center}
        \includegraphics[width=2.0\columnwidth]{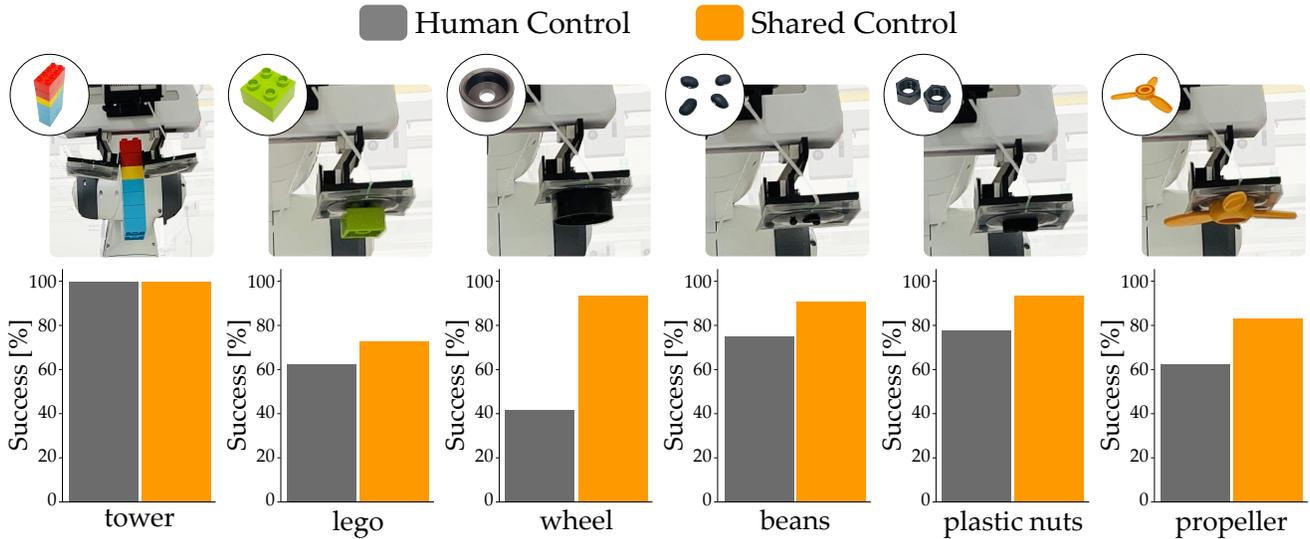}
        \caption{RISO gripper picking up example objects from our user study. The first column demonstrates the use of the rigid mechanism, while columns $2$-$6$ highlight the soft adhesives. (Top Row) Images of the RISO gripper holding items. (Bottom Row) $10$ participants controlled the robot to try and grasp these items using \textbf{Human Control} and \textbf{Shared Control}. We plot the average success per item across all users (i.e., the percentage of time that users were able to pick up the target item). Success was comparable for \textbf{Human} and \textbf{Shared} when grasping an object using the rigid gripper (column $1$). However, \textbf{Shared} outperformed \textbf{Human} when participants attempted to leverage the soft adhesives (columns $2$-$6$).}
        \label{fig:grasp}
    \end{center}
    \vspace{-2.0em}
\end{figure*}

\section{User Study}

We conducted an in-person user study to evaluate the performance of the RISO gripper. During this study participants controlled a robot arm to pick up, move, and release a diverse set of objects (see Table~\ref{tab:results} and \fig{grasp}). We compared the gripper's performance when users completely controlled the robot, and when the robot applied our proposed approach to partially automate the grasping process. Our results suggest that humans and robots can leverage the RISO gripper to manipulate objects using either the rigid parallel mechanism or the soft switchable adhesive. We include videos from this user study here: \href{https://youtu.be/5uLUkBYcnwg}{https://youtu.be/5uLUkBYcnwg}

\p{Experimental Setup}
For this study we used a $7$-DoF robot arm (Franka Emika) with an industry-standard two-fingered rigid gripper \cite{frankagripper}. We mounted soft adhesives to $3$D printed parts at the bottom of both rigid fingers (see \fig{front}). The resulting RISO gripper could pick up objects using either the standard rigid mechanism (i.e., squeezing an object between both fingers) or using a switchable soft adhesive (i.e., pushing one finger into contact with the object, and then activating the adhesive). Participants teleoperated this robot arm and RISO gripper using a hand-held joystick (SteelSeries Stratus Duo). Using a joystick to teleoperate the robot is consistent with recent work on assistive robot arms \cite{beaudoin2019long, jain2016grasp, losey2022learning, jain2019probabilistic, javdani2018shared}.

\p{Objects} During the user study the robot and RISO gripper needed to pick up, move, and release a total of $15$ household objects. The majority of these items ($9/15$) are from the YCB dataset for robotic manipulation research \cite{calli2017yale}. We then added $6$ more objects: these included foods (black beans, M\&M's, and a chocolate syrup bottle) and everyday items (metal nuts, glue bottle, and a fidget spinner). Out of the $15$ total items, we instructed participants to grasp the $3$ largest and heaviest using the rigid gripper (chocolate syrup, Lego tower, and glue bottle). Users manipulated the remaining $12$ objects using soft adhesives. We list all $15$ objects in Table~\ref{tab:results}.

\p{Task} We randomly placed all $15$ household items on a table in front of the robot arm. Participants were instructed to clear the table, one item at a time. Users teleoperated the robot to grasp an object of their choice, carry that object over to a bin, release the item, and then move on to the next object. The robot observed the position of objects in real-time using a depth camera (Intel RealSense) mounted on its arm.

\p{Independent Variables} Participants interacted with two different RISO control algorithms:
\begin{itemize}
    \item \textbf{Human}, a control mode where users directly teleoperated the arm and RISO gripper at every timestep.
    \item \textbf{Shared}, our proposed approach from Section~\ref{sec:control}.
\end{itemize}
In both \textbf{Human} and \textbf{Shared} the user had \textit{direct control} over the force applied by the rigid gripper and the input pressure of the soft adhesives. For instance, users pressed the right and left bumpers on the joystick to increment $\Delta P$ and change the pressure in \eq{C1}. The difference between \textbf{Human} and \textbf{Shared} was in the movement of the robot arm: \textbf{Shared} \textit{partially automated} the arm's motion to align the RISO gripper with the human's target object. When using \textbf{Human} the participant had to directly control the velocity of the robot's end-effector throughout the process of reaching for, aligning with, and then carrying the object. In \textbf{Shared} the robot attempted to automate the process of reaching for the object and aligning the gripper, and participants used the joystick to make minor adjustments as necessary. \textbf{Shared} also helped maintain the RISO gripper's alignment and applied force during the grasping process.

\p{Dependent Variables}
We recorded each interaction to determine how effective the RISO gripper was with \textbf{Human} and \textbf{Shared} control. First, we measured the percentage of objects that the robot successfully picked up, moved, and released in the bin (\textit{Success}). We next measured the average amount of time participants spent per object (\textit{Grasp Time}). For this metric --- and the following metrics --- we only considered parts of the task where the robot arm was directly above the table (i.e., the segments of the task when the robot could interact with objects). To compute \textit{Grasp Time} we divided the total interaction time by $15$. We similarly recorded \textit{Grasp Distance}, the average distance that the robot's end-effector traveled per object, and \textit{Input Time}, the average time that participants spent providing joystick inputs. Lower values of \textit{Grasp Time} and \textit{Input Time} show that participants completed the task more quickly and used less time controlling the robot, while lower \textit{Grasp Distance} indicates the robot took more efficient paths to and from the objects.

\p{Participants and Procedure} We recruited $10$ participants from the Virginia Tech community ($2$ female, average age $25.3\pm 4.7$ years). Participants provided informed written consent under Virginia Tech IRB $\#22$-$308$. We utilized a within-subject design: each participant completed the task twice, clearing the table of all items once with \textbf{Human} and once with \textbf{Shared}. We balanced the order of presentation, so that half of the participants started with \textbf{Shared} and the other half started with \textbf{Human}. Before the start of the experiment users were given $5$ minutes to practice manipulating objects and familiarizing themselves with the joystick interface, robot arm, and RISO gripper.

\begin{figure*}[t]
    \begin{center}
        \includegraphics[width=2.0\columnwidth]{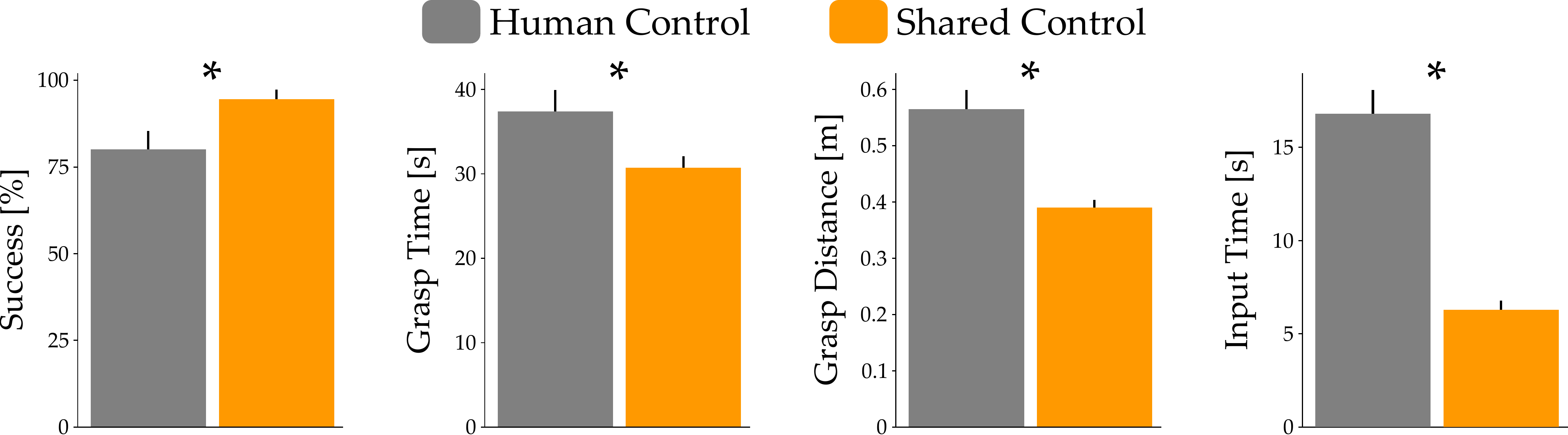}
        \caption{Objective results from our user study. During this study we placed $15$ household items on a table, and participants controlled the robot arm and RISO gripper to pick up each object and drop it in a bin. We compared two control approaches: \textbf{Human Control}, where the user teleoperated each step of the process, and \textbf{Shared Control}, where the autonomous robot helped to align the RISO gripper for the human's target object and grasp type. \textbf{Shared Control} led to a higher overall success rate (\textit{Success}), less time spent picking up objects (\textit{Grasp Time}), less robot motion (\textit{Grasp Distance}), and users spent less time interacting with the joystick (\textit{Input Time}). Error bars show standard error and an $*$ denotes statistical significance ($p<.05$).}
        \label{fig:results}
    \end{center}
    \vspace{-1.5em}
\end{figure*}

\p{Hypothesis} We hypothesized that:
\begin{quote}
\p{H1} \textit{RISO grippers will pick up, move, and release more objects with shared control.}
\end{quote}
\begin{quote}
\p{H2} \textit{Users will complete the task more quickly and efficiently using shared control.}
\end{quote}

\p{Results} Examples of the RISO holding objects are shown in \fig{grasp}, and we list the success rate for each individual object in \fig{grasp} and Table~\ref{tab:results}. The average grasp success and other objective metrics are displayed in \fig{results}.

Overall, we found that participants successfully picked up, moved, and released more objects when the robot partially automated the grasping process. Paired t-tests reveal that the average success across all $15$ items is higher with \textbf{Shared} as compared to \textbf{Human} ($t(9) = -3.454$, $p<.01$). We highlight that both \textbf{Human} and \textbf{Shared} performed the same with the rigid gripper, but differences emerged when leveraging the soft adhesives. With \textbf{Human} participants struggled to pick up objects that were almost as large as the adhesive (e.g., the big toy wheel) and groups of smaller objects (e.g., the beans, M\&M's, or nuts). Here \textbf{Shared} outperformed \textbf{Human} because (a) the assistive robot better aligned its adhesive with the target, (b) the assistive robot held that alignment while the human set the input pressure of the RISO gripper, and (c) the assistive robot maintained a constant contact force as the adhesive pressure changed.

We next focused on how efficiently participants completed the task. Referring to \fig{results}, we found that users took less time to pick up, move, and release an object using \textbf{Shared} as compared to \textbf{Human} ($t(9) = 2.623$, $p<.05$). Users also performed this manipulation using more efficient trajectories: with \textbf{Human} the participants often had to reverse course or change the gripper position, causing the RISO gripper to travel a longer distance with \textbf{Human} ($t(9) = 5.636$, $p<.05$). Finally, users spent less time providing controller inputs under \textbf{Shared} ($t(9) = 9.279$, $p<.05$). This result matches our expectations because with \textbf{Human} the user must control the robot throughout the entire task, while with \textbf{Shared} the robot infers the human's intent and takes autonomous actions to try and reduce the human's workload.

Our collective results from this study support hypotheses \textbf{H1} and \textbf{H2}. Both the human and robot should have an active role when controlling a RISO gripper; if we share control between agents, the RISO gripper can successfully grasp a variety of household items.

\begin{table}[t]
\caption{Percentage of interactions where participants successfully picked up, moved, and released each object in the user study. For the top three objects we instructed participants to leverage the rigid gripper; participants used the soft adhesive for the remaining items.}
\label{tab:results}
\resizebox{\columnwidth}{!}{%
\begin{tabular}{@{}llcc@{}}
\toprule
 \textbf{Grasp}&
  \multicolumn{1}{c}{\textbf{Objects}} &
  \begin{tabular}[c]{@{}c@{}}\textbf{Human}\\  Success [\%]\end{tabular} &
  \multicolumn{1}{c}{\begin{tabular}[c]{@{}c@{}}\textbf{Shared} \\ Success [\%]\end{tabular}} \\ \midrule
\multicolumn{1}{l|}{\multirow{3}{*}{\rotatebox[origin=c]{90}{\begin{tabular}[c]{@{}c@{}}\textbf{Rigid} \\ \textbf{Gripper}\end{tabular}}}} &
  Chocolate Syrup &
  $100\%$ &
  $100\%$ \\
\multicolumn{1}{l|}{} & Glue Bottle            & $100\%$   & $100\%$   \\
\multicolumn{1}{l|}{} & Lego Tower             & $100\%$   & $100\%$   \\ \midrule
\multicolumn{1}{l|}{\multirow{12}{*}{\rotatebox[origin=c]{90}{\begin{tabular}[c]{@{}c@{}}\textbf{Soft} \\ \textbf{Adhesive}\end{tabular}}}} &
Beans                  & $72.5\%$ & $87.5\%$ \\
\multicolumn{1}{l|}{} & Dice                   & $100\%$   & $100\%$   \\
\multicolumn{1}{l|}{} & Fidget Spinner         & $100\%$   & $100\%$   \\
\multicolumn{1}{l|}{} & Lego Block             & $60\%$    & $70\%$    \\
\multicolumn{1}{l|}{} & Metal Nuts             & $80\%$    & $100\%$   \\
\multicolumn{1}{l|}{} & M\&M's                 & $77.5\%$ & $100\%$ \\
\multicolumn{1}{l|}{} & Plastic Nuts           & $75\%$  & $90\%$    \\
\multicolumn{1}{l|}{} & Toy Propeller           &
  $60\%$ &
  $80\%$ \\
\multicolumn{1}{l|}{} & Toy Wheel (Big)   & $40\%$    & $90\%$    \\
\multicolumn{1}{l|}{} & Toy Wheel (Small) & $90\%$    & $100\%$   \\
\multicolumn{1}{l|}{} & Washer                 & $60\%$    & $90\%$    \\
\multicolumn{1}{l|}{} & Wooden Block           & $90\%$    & $100\%$   \\ \bottomrule
\end{tabular}%
}
\end{table}

\section{Conclusion}

In this paper we introduced RIgid-SOft (RISO) grippers for assistive and industrial robot arms. We first formed RISO grippers by attaching a novel class of soft, switchable adhesives to the fingers of rigid grippers. By adjusting the pinching force of the rigid fingers RISO grippers can hold large and heavy items, and by changing the input pressure of the soft adhesives RISO grippers can carry numerous small objects. We next formulated a shared control approach to help human operators leverage RISO grippers. This approach partially automates the motion of the robot arm to align the RISO gripper with the object and grasp that the robot infers from the human. Finally, we conducted a user study to demonstrate that RISO grippers can pick up a variety of household items. Our results suggest that integrating rigid and soft grippers into RISOs and combining this architecture with shared control leads to higher grasping success and more efficient interaction.

\p{Limitations and Future Work} While performing experiments with the RISO gripper we observed that --- even though it can pick up large objects using the parallel grasping mechanism --- there are some limitations to the objects that it can grasp using the switchable adhesive. For example, due to the single discrete switchable adhesive, objects larger than the adhesive surface can be challenging to manipulate. This may be alleviated by creating an array of switchable adhesives or modifying the amount of switchable adhesive on the rigid gripping surface.

Next, the contact area between the adhesive gripper and target object must be controlled. At smaller length scales this means creating contact area on rough surfaces, which can be accomplished with softer adhesive materials and further development of the pneumatic control. At larger scales, parallelism between the gripper and the target surface must be considered. This is particularly important for gripping flat objects, where large misalignment can lead to decreased grasping success. We envision solutions that (a) use control strategies for the robot arm to \textit{actively} minimize misalignment, and (b) incorporate compliant linkages that \textit{passively} align the soft adhesive and target object. 

Finally, fully automated robot control could provide capabilities beyond the human or shared control strategies. This area of future work will require analysis of the objects to be grasped and improved path and grasp planning. For objects, classification into groups based on size, texture, and stiffness could allow for informed grasp strategies. For path planning, consideration of object size will be important and grasp planning will allow for effective and efficient manipulation of a greater range of objects moving forward.



\balance
\bibliographystyle{IEEEtran}
\bibliography{references}
\end{document}